\PassOptionsToPackage{numbers}{natbib}
\documentclass{article}
\usepackage[final]{neurips_2025_ml4ps}

\usepackage[utf8]{inputenc}
\usepackage[T1]{fontenc}
\usepackage{microtype}
\usepackage{xcolor}
\usepackage{ifthen}
\usepackage{url}
\usepackage{booktabs}
\usepackage{nicefrac}
\usepackage{amsmath, amsfonts, amssymb}
\usepackage{mathtools}
\usepackage{bm}
\usepackage{bbm}
\usepackage{siunitx}
\usepackage{float}
\usepackage{algorithm}
\usepackage{algorithmic}
\usepackage{graphicx}
\usepackage{subcaption}
\usepackage{adjustbox}
\usepackage{tabularx}
\usepackage{placeins}
\usepackage{tikz}
\usepackage{multirow}
\newcolumntype{Y}{>{\raggedright\arraybackslash}X}

\usetikzlibrary{arrows.meta,positioning,calc,fit,shapes.geometric}
\definecolor{featcol}{RGB}{224,243,255}
\definecolor{modcol}{RGB}{232,246,231}
\definecolor{opcol}{RGB}{245,239,230}
\definecolor{updcol}{RGB}{255,236,204}
\tikzset{
  >={Latex[length=2.1mm]},
  box/.style={draw, rounded corners=2pt, line width=0.8pt, inner sep=4pt, align=center, fill=#1},
  feat/.style={box=featcol},
  module/.style={box=modcol},
  op/.style={box=opcol},
  update/.style={box=updcol},
  flow/.style={->, line width=0.85pt, rounded corners=3pt, shorten >=1pt, shorten <=1pt},
  small/.style={font=\small},
  tiny/.style={font=\footnotesize}
}

\usepackage[hidelinks]{hyperref}
\usepackage[nameinlink,noabbrev,capitalise]{cleveref}

\newcommand{\E}{\mathbb{E}}
\newcommand{\R}{\mathbb{R}}

\newcommand{\RJCP}{\ensuremath{\mathsf{RJCP}}}
\newcommand{\DIPG}{\textsc{D-IPG}}
\newcommand{\xGD}{\textsc{x-GD}}
\newcommand{\GN}{\textsc{GN/LM}}

\title{Deceptron: Learned Local Inverses for Fast and Stable Physics Inversion\thanks{
The naming ``Deceptron'' reflects its contrast with the forward-mapping perceptron.}}

\author{%
  Aaditya L. Kachhadiya\\
  Independent Researcher\\
  Surat, India\\
  \texttt{kachhadiyaaaditya@gmail.com} \\
}

\begin{document}

\maketitle

\begin{abstract}
Inverse problems in the physical sciences are often ill-conditioned in input space, making progress step-size sensitive. We propose the \emph{Deceptron}, a lightweight bidirectional module that learns a local inverse of a differentiable forward surrogate. Training combines a supervised fit, forward--reverse consistency, a lightweight spectral penalty, a soft bias tie, and a \emph{Jacobian Composition Penalty} (JCP) that encourages $J_g(f(x))\,J_f(x)\!\approx\!I$ via JVP/VJP probes. At solve time, \DIPG{}\ (Deceptron Inverse-Preconditioned Gradient) takes a descent step in output space, pulls it back through $g$, and projects under the same backtracking and stopping rules as baselines. On Heat-1D initial-condition recovery and a Damped Oscillator inverse problem, \DIPG{}\ reaches a fixed normalized tolerance with $\sim$20$\times$ fewer iterations on Heat and $\sim$2--3$\times$ fewer on Oscillator than projected gradient, competitive in iterations and cost with Gauss--Newton. Diagnostics show JCP reduces a measured composition error and tracks iteration gains. We also preview a single-scale 2D instantiation, \emph{DeceptronNet (v0)}, that learns few-step corrections under a strict fairness protocol and exhibits notably fast convergence.
\end{abstract}

\section{Introduction}
Recovering unknown inputs or parameters from indirect, noisy measurements is central to PDE inversion, system identification, and imaging. A common approach minimizes a data misfit in input space with projections enforcing physics constraints. Such objectives are often ill-conditioned; gradients are poorly scaled and many iterations are required. We propose the \textbf{Deceptron}, a simple alternative: learn a local inverse of a differentiable forward surrogate and use it to precondition inverse updates. A small bidirectional module parameterizes a forward map $f$ and a reverse map $g$; training stabilizes $g$ as a local inverse via a JVP/VJP-based penalty on $J_g(f(x))\,J_f(x)\!\approx\!I$. At inference, \DIPG\ takes a residual step in output space, pulls it back through $g$, then projects with the same Armijo backtracking used by baselines \cite{Armijo1966}. This preserves a standard projected loop and changes only the update direction. Beyond this single-module preconditioning, we outline a broader agenda: a \emph{DeceptronNet} family tailored to inverse problems. As a first step, we introduce a single-scale 2D unrolled variant (v0) that maps nominal residual features to image-space corrections in a few learned steps. We compare Deceptron primarily against Gauss--Newton (GN) and gradient descent in the $x$-space (\xGD{}). Here, \xGD{} denotes plain gradient descent updates directly on $x$, 
while \DIPG{} refers to updates carried out in the data space $y$ and then pulled back through $g$.

Related efforts include physics-informed training \cite{Raissi2019PINNs,Karniadakis2021Physics}, learned unrolling and proximal priors \cite{Gregor2010LISTA,Venkatakrishnan2013Plug,Romano2017RED}. Classical inverse methods like Gauss--Newton/LM \cite{Levenberg1944,Marquardt1963} guide our comparison; for Hessian--vector products we follow Pearlmutter \cite{Pearlmutter1994}.

\section{Method}

Let $x\!\in\!\R^{d_\text{in}}$ and $y\!\in\!\R^{d_\text{out}}$. The Deceptron defines
\[
f_W(x)=\sigma(Wx+b), \qquad g_V(y)=\tilde\sigma(Vy+c),
\]
where $W,b,V,c$ are learned parameters and $\sigma,\tilde\sigma$ are lightweight activation functions (e.g., leaky). The matrices $V$ and $W^\top$ are not tied so that $g$ can act as a local inverse even when $W$ is non-orthogonal. Stabilization terms include $\lVert W^\top W-I\rVert_F^2$, a soft bias tie $\lVert b+c\rVert_2^2$, and optionally $\lVert VW-I\rVert_F^2$.

Given pairs $(x,y^\ast)$ from a differentiable surrogate, the loss is
\begin{align}
\mathcal{L} &=
\lambda_\text{task}\lVert f_W(x)-y^\ast\rVert^2
+\lambda_\text{rec}\lVert g_V(f_W(x))-x\rVert^2
+\lambda_\text{cyc}\lVert f_W(g_V(\tilde y))-\tilde y\rVert^2 \nonumber\\
&\quad+\beta_\text{spec}\lVert W^\top W-I\rVert_F^2
+\lambda_\text{tie}\lVert b+c\rVert_2^2
+\lambda_\text{comp}\lVert VW-I\rVert_F^2 \nonumber\\
&\quad+\lambda_\text{JCP}\,\mathbb{E}_{\xi}\lVert J_g(f_W(x))J_f(x)\xi - \xi \rVert^2. \label{eq:loss}
\end{align}
Here $\tilde y$ denotes measurement-space samples such as $f_W(x)$ or noised variants. With probes $\mathbb{E}[\xi\xi^\top]=I$ (Rademacher or Gaussian, one to four per batch), the identity $\mathbb{E}\lVert(A-I)\xi\rVert^2=\lVert A-I\rVert_F^2$ ensures that the JCP term estimates $\lVert J_g(f(x))J_f(x)-I\rVert_F^2$ using only a few JVP/VJP products.

We minimize $\Phi(x)=\tfrac12\lVert f_W(x)-y^\ast\rVert^2$ in normalized output space. At iteration $t$, let $y_t=f_W(x_t)$ and $r_t=y_t-y^\ast$. The update is
\[
y_{t+1}^{\text{prop}}=y_t-\alpha r_t, \qquad
x_{t+1}^{\text{prop}}=g_V(y_{t+1}^{\text{prop}}), \qquad
x_{t+1}=\Pi_{\mathcal{C}}\big((1-\rho)x_t+\rho\,x_{t+1}^{\text{prop}}\big),
\]
accepted under the shared Armijo and projection rule. To first order,
$g(y_t-\alpha r_t)\approx x_t-\alpha\,J_g(f(x_t))r_t$. If $J_f(x_t)$ has full column rank and
$J_g(f(x_t))\approx J_f(x_t)^+=(J^\top J)^{-1}J^\top$,
then
\DIPG\ matches Gauss–Newton up to the scalar step size $\alpha$; as $\lVert J_g(f(x))J_f(x)-I\rVert\!\to\!0$, the updates converge to Gauss–Newton scaled by $\alpha$ \cite{Levenberg1944,Marquardt1963}. Limitations include locality and surrogate fidelity (see Appendix~\ref{app:limits}).

We monitor the runtime Jacobian composition error
\RJCP$(x)=\mathbb{E}_{\xi}\lVert J_g(f(x))J_f(x)\xi-\xi\rVert^2$,
an unbiased estimator of $\lVert J_g(f(x))J_f(x)-I\rVert_F^2$ obtained via Hutchinson’s identity.
\RJCP$(x)=0$ if and only if $J_g(f(x))J_f(x)=I$, meaning $g$ is a local left inverse at $x$.
Lower values of $\RJCP$ empirically correlate with fewer iterations (Fig.~\ref{fig:jcp_rjcp_bar}).

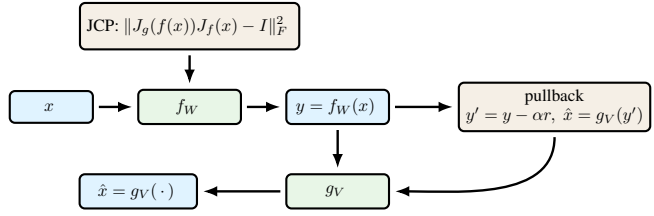
\begin{figure}[t]
\centering
\begin{tikzpicture}[
  scale=0.65, every node/.style={transform shape},
  >={Latex[length=1.9mm]},
  flow/.style={->, line width=0.9pt, rounded corners=2pt, shorten >=2pt, shorten <=2pt}
]
\node[feat,   minimum width=1.7cm, minimum height=6.5mm] (x)   { $x$ };
\node[module, minimum width=2.1cm, minimum height=7mm, right=9mm of x] (fw) { $f_W$ };
\node[feat,   minimum width=2.1cm, minimum height=7mm, right=9mm of fw] (y)   { $y=f_W(x)$ };
\node[module, minimum width=2.1cm, minimum height=7mm, below=10mm of y] (gv)  { $g_V$ };
\node[feat,   minimum width=2.4cm, minimum height=7mm, left=18mm of gv] (xh)  { $\hat{x}=g_V(\,\cdot\,)$ };
\node[op,     minimum width=3.3cm, minimum height=9.5mm, above=8mm of fw] (jcp)
  { JCP: $\lVert J_g(f(x))J_f(x)-I\rVert_F^2$ };
\node[op,     minimum width=3.6cm, minimum height=9.5mm, right=14mm of y] (pull)
  { pullback\\ $y' = y-\alpha r,\;\hat{x}=g_V(y')$ };
\draw[flow] (x.east) -- (fw.west);
\draw[flow] (fw.east) -- (y.west);
\draw[flow] (y.south) -- (gv.north);
\draw[flow] (gv.west) -- (xh.east);
\draw[flow] (y.east) -- (pull.west);
\draw[flow] (pull.south) to[out=270,in=0] (gv.east);
\draw[flow] (jcp.south) -- (fw.north);
\end{tikzpicture}
\caption{Deceptron: forward $f_W$ and reverse $g$ (instantiated as learned $g_V$ by default, or $g_{W^\top}$ if tied) with JCP; inference pulls output-space residuals back through $g$.}
\label{fig:deceptron-structure}
\end{figure}

All algorithms are listed in the Appendix.\footnote{See pseudocode in Algorithms~\ref{alg:dipg}--\ref{alg:DNet}.}

\section{Experiments}

We evaluate the Deceptron inverse-preconditioned gradient (\DIPG{}) on two standard inverse problems: Heat-1D initial-condition recovery and Damped Oscillator parameter and initial-condition estimation. Outputs are z-scored, and all algorithms share the same normalized loss space ($\varepsilon\!=\!0.30$). This keeps the line search and stopping policy comparable across methods even when the raw scales of $y$ differ by task. Final RMSE values in \cref{tab:fair,tab:heat_robust,tab:osc_robust} are reported in unnormalized units to reflect physical error.

All solvers follow an identical fairness protocol. The same projector $\Pi_{\mathcal{C}}$ and Armijo rule with $c=10^{-4}$ (up to eight halvings) are used for all methods \cite{Armijo1966}. Relaxation is fixed to $\rho=0.4$, and the same initial step size $1.0$ is used for each optimizer (\xGD{} $\eta$, \DIPG{} $\alpha$, \GN{} $\alpha$). The maximum iteration count is 200. Heat-1D begins from a zero initial condition, while the oscillator starts from a mid-range parameter vector. Backtracking evaluations of $f$ are shared to isolate only the effect of update direction. No proximal or smoothing heuristics are used, and all runs are deterministic with fixed seeds.

\begin{figure}[t]
  \centering
  \begin{subfigure}{0.26\linewidth}
    \includegraphics[width=\linewidth]{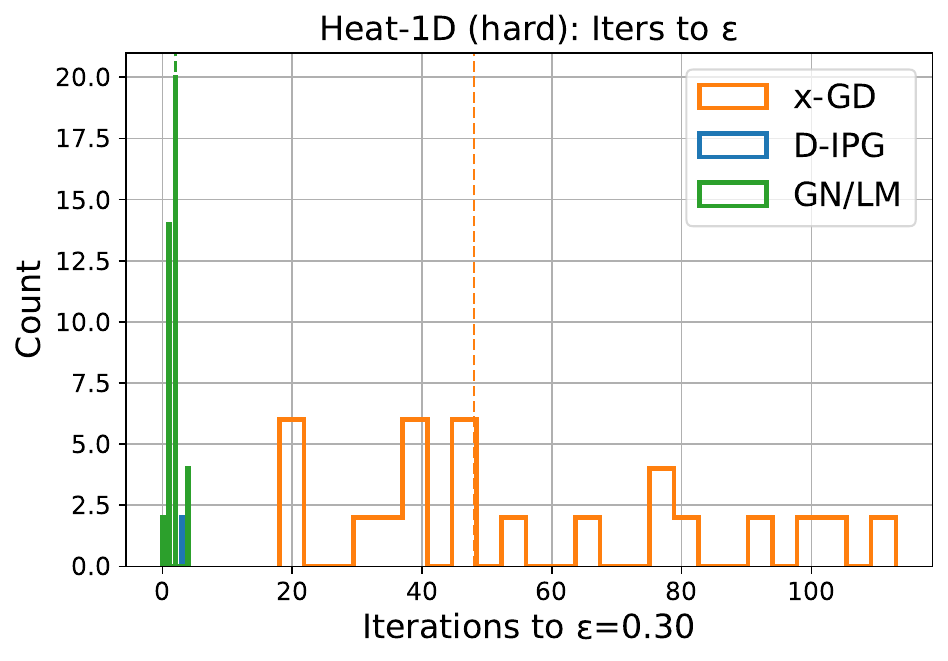}
    \caption{Heat-1D iterations}
    \label{fig:heat_hist}
  \end{subfigure}\hfill
  \begin{subfigure}{0.25\linewidth}
    \includegraphics[width=\linewidth]{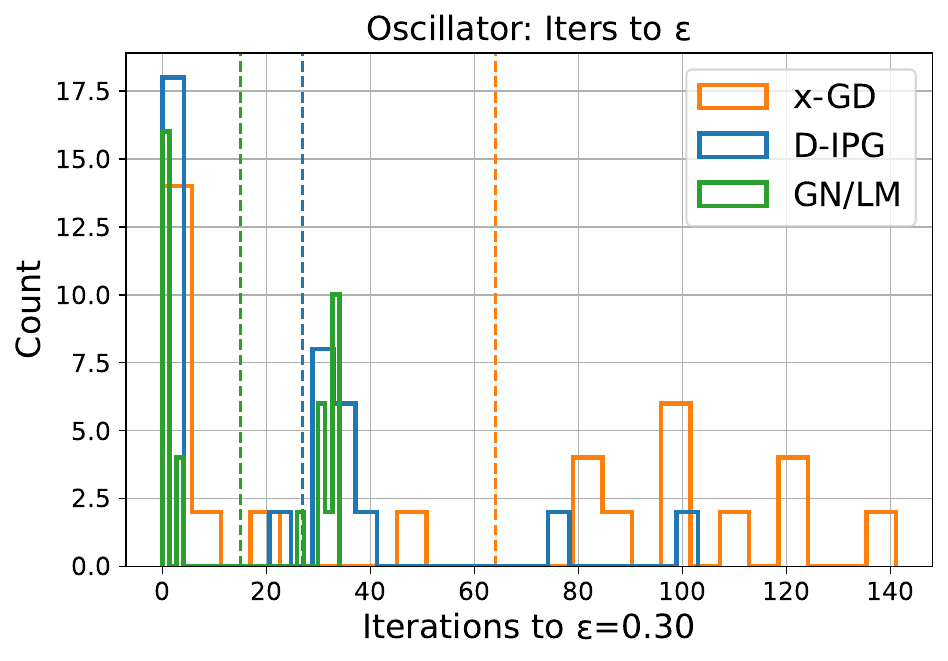}
    \caption{Oscillator iterations}
    \label{fig:osc_hist}
  \end{subfigure}\hfill
  \begin{subfigure}{0.45\linewidth}
    \includegraphics[width=\linewidth]{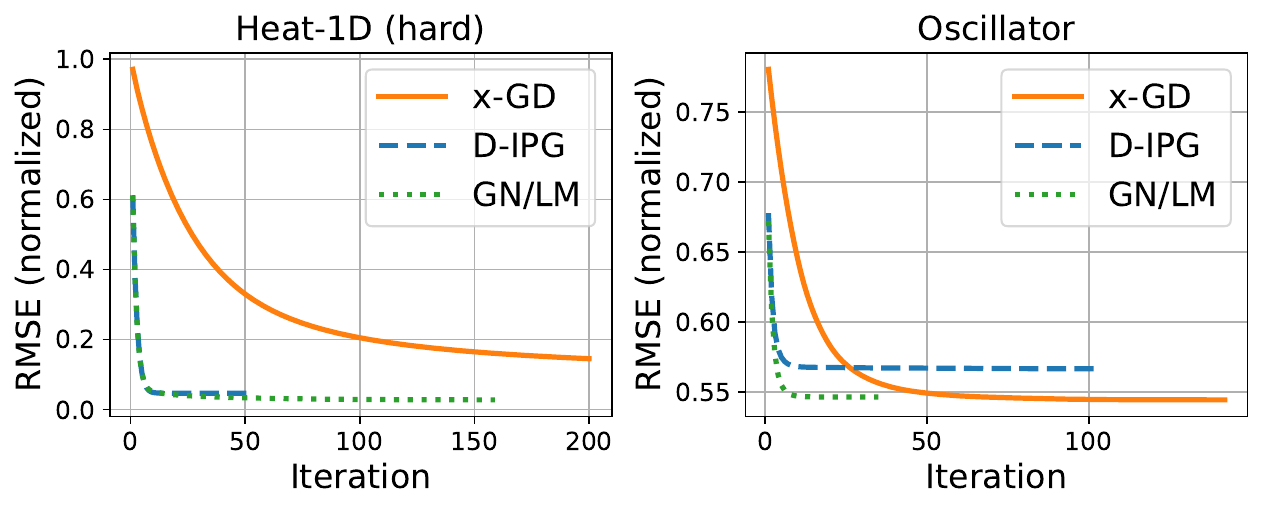}
    \caption{Trajectory curves}
    \label{fig:rmse_traj}
  \end{subfigure}
  \caption{Iteration distributions and convergence trajectories across problems. The right panel includes both Heat-1D and Oscillator RMSE curves (mean normalized RMSE).}
  \label{fig:main_combo}
\end{figure}

\cref{fig:main_combo} compares iteration counts and normalized trajectories under the shared policy. On Heat-1D, \DIPG{} and \GN{} concentrate at very low iteration counts while \xGD{} is widely spread, indicating sensitivity to poor conditioning in $x$-space. The trajectory panel shows that all methods eventually reduce the normalized residuals, but the preconditioned directions of \DIPG{} and the second-order curvature of \GN{} reach the tolerance in a few steps. On the oscillator problem, \GN{} has a slight iteration edge over \DIPG{}, which is consistent with its access to explicit Hessian information, yet the per-iteration cost of \GN{} is higher due to inner linear solves. The separation between methods therefore reflects both direction quality and compute per step.

\begin{minipage}{\linewidth}\centering
\footnotesize\setlength{\tabcolsep}{3.5pt}
\captionof{table}{Iterations-to-$\varepsilon$ (mean$\pm$std), final RMSE, and acceptance rate (acc).}
\label{tab:fair}
\begin{adjustbox}{max width=\linewidth}
\begin{tabular}{lccccccccc}
\toprule
& \multicolumn{3}{c}\xGD{} & \multicolumn{3}{c}\DIPG{} & \multicolumn{3}{c}\GN{} \\
\cmidrule(lr){2-4}\cmidrule(lr){5-7}\cmidrule(lr){8-10}
Setting & it & RMSE & acc & it & RMSE & acc & it & RMSE & acc \\
\midrule
Heat-1D (hard) & $58.2\!\pm\!28.9$ & $0.045$ & $1.00$ & $2.8\!\pm\!1.0$ & $0.010$ & $0.58$ & $2.8\!\pm\!0.9$ & $0.009$ & $0.97$ \\
Oscillator     & $58.2\!\pm\!52.1$ & $0.356$ & $1.00$ & $24.6\!\pm\!27.2$ & $0.368$ & $0.64$ & $17.3\!\pm\!15.7$ & $0.353$ & $0.69$ \\
\bottomrule
\end{tabular}
\end{adjustbox}
\end{minipage}

\cref{tab:fair} summarizes averages over trials (unnormalized final RMSE). The acceptance rate is lower for \DIPG{} on Heat-1D because its proposals are larger; Armijo reduces the step a few times before acceptance, which is expected under stronger preconditioning and does not indicate instability. The final RMSE values in original units are comparable between \DIPG{} and \GN{} and significantly better than \xGD{} on Heat-1D. On the oscillator problem, \DIPG{} has more variability in iteration counts, but still outperforms \xGD{} and approaches \GN{}.

\begin{minipage}{\linewidth}\centering
\footnotesize\setlength{\tabcolsep}{4.5pt}
\captionof{table}{Heat-1D ($\varepsilon{=}0.30$). Median [IQR] iters, success, ms/iter, and mean time-to-$\varepsilon$.}
\label{tab:heat_robust}
\begin{adjustbox}{max width=\linewidth}
\begin{tabular}{lcccc}
\toprule
Method & iters [IQR] & Success & ms/iter & Time (s) \\
\midrule
\xGD{}  & $49.0\,[38.2,\,80.0]$ & $1.00$ & $0.43$ & $0.026$ \\
\DIPG{} & $3.0\,[2.0,\,3.0]$   & $1.00$ & $0.51$ & $0.001$ \\
\GN{}   & $3.0\,[2.0,\,3.0]$   & $1.00$ & $3.82$ & $0.011$ \\
\bottomrule
\end{tabular}
\end{adjustbox}
\end{minipage}

\begin{minipage}{\linewidth}\centering
\footnotesize\setlength{\tabcolsep}{4.5pt}
\captionof{table}{Oscillator ($\varepsilon{=}0.30$). Median [IQR] iters, success, ms/iter, and mean time-to-$\varepsilon$.}
\label{tab:osc_robust}
\begin{adjustbox}{max width=\linewidth}
\begin{tabular}{lcccc}
\toprule
Method & iters [IQR] & Success & ms/iter & Time (s) \\
\midrule
\xGD{}  & $65.0\,[1.0,\,104.5]$ & $0.50$ & $0.45$ & $0.004$ \\
\DIPG{} & $28.0\,[1.0,\,34.0]$  & $0.45$ & $1.28$ & $0.001$ \\
\GN{}   & $16.5\,[1.0,\,33.2]$  & $0.50$ & $4.22$ & $0.007$ \\
\bottomrule
\end{tabular}
\end{adjustbox}
\end{minipage}

The median summaries make the tradeoff clear. \DIPG{} matches \GN{} in iteration counts on Heat-1D but has much lighter iterations, resulting in shorter mean time-to-tolerance. On the oscillator, \GN{} reduces iteration counts further but pays a higher cost per step, whereas \DIPG{} retains a good balance between direction quality and compute. This is consistent with a Gauss–Newton–like direction from \DIPG{} when $J_g(f)J_f$ is close to identity, but without solving linear systems every iteration.

\begin{figure}[t]
  \centering
  \begin{subfigure}{0.25\linewidth}
    \includegraphics[width=\linewidth]{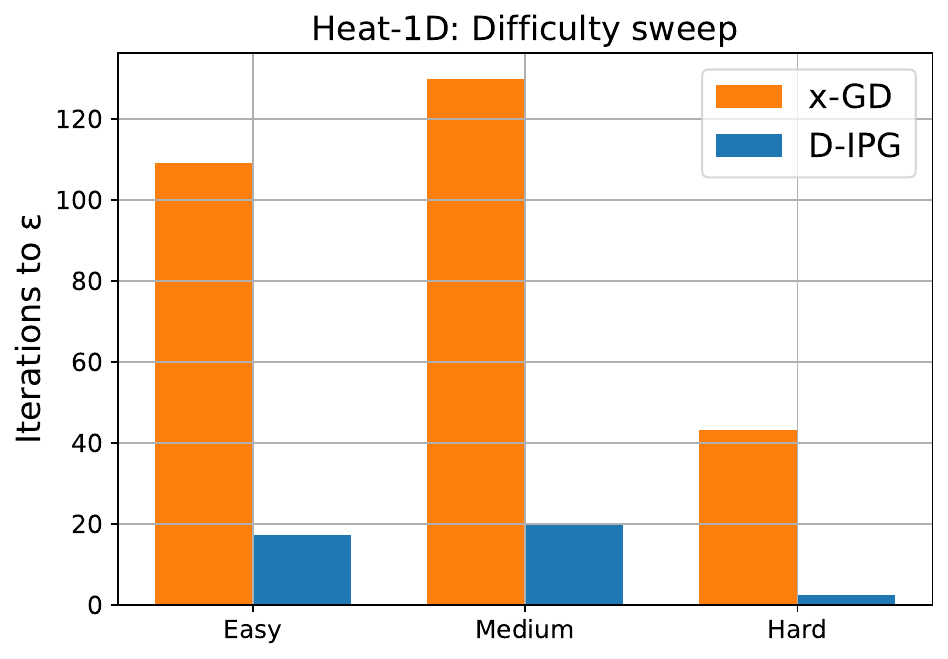}
    \caption{Heat-1D sweep}
    \label{fig:heat_sweep}
  \end{subfigure}\hfill
  \begin{subfigure}{0.25\linewidth}
    \includegraphics[width=\linewidth]{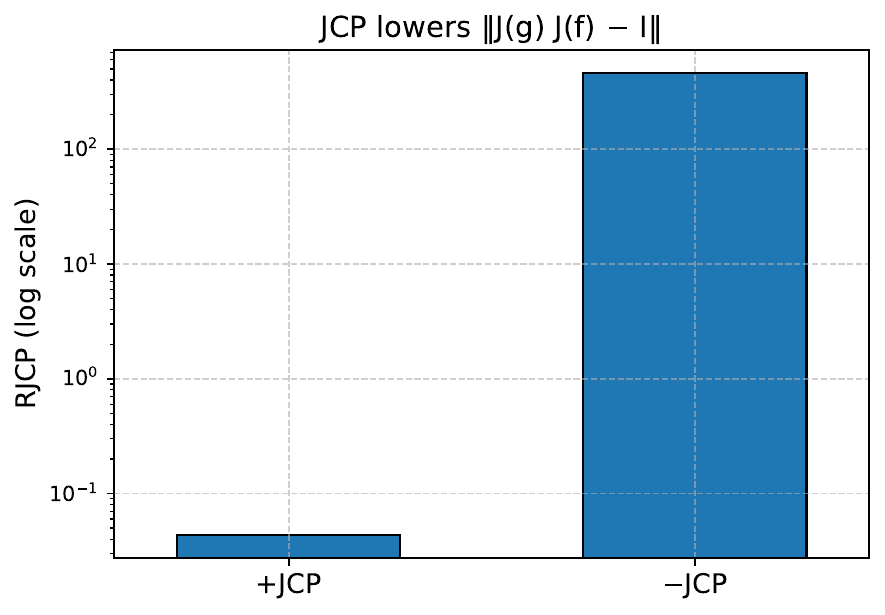}
    \caption{$\RJCP$ consistency}
    \label{fig:jcp_rjcp_bar}
  \end{subfigure}\hfill
  \begin{subfigure}{0.25\linewidth}
    \includegraphics[width=\linewidth]{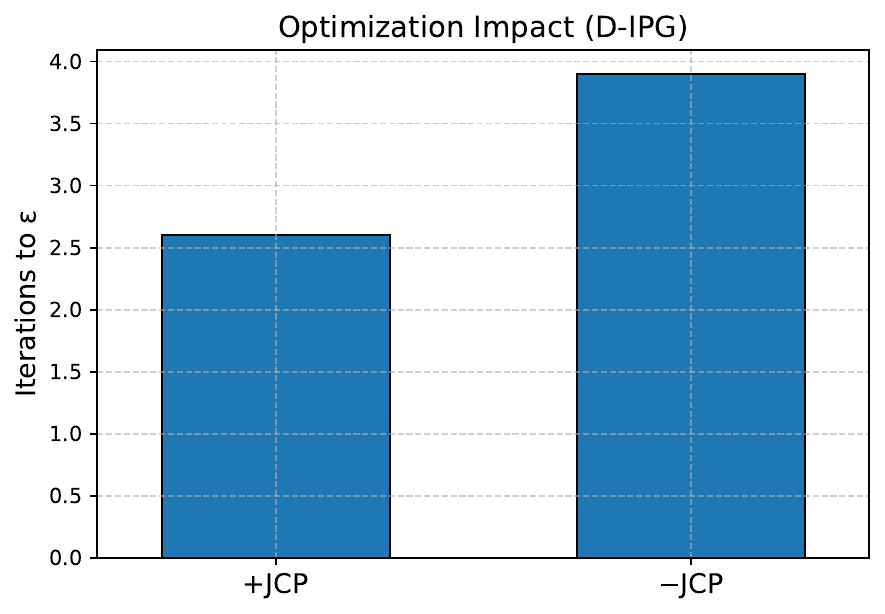}
    \caption{Optimization impact}
    \label{fig:jcp_iters_bar}
  \end{subfigure}\hfill
  \begin{subfigure}{0.25\linewidth}
    \includegraphics[width=\linewidth]{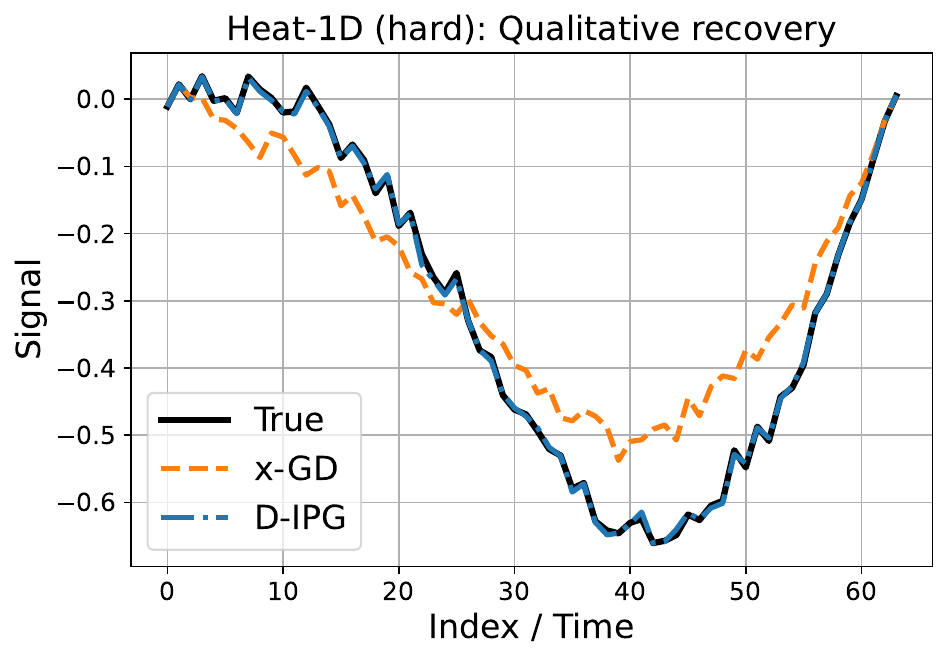}
    \caption{Qualitative recovery}
    \label{fig:qual_recov}
  \end{subfigure}
  \caption{Scaling and ablations. \DIPG{} remains stable under increasing Heat-1D difficulty, JCP lowers composition error and iteration count, and final reconstructions confirm accuracy.}
  \label{fig:ablations}
\end{figure}

We next study how convergence changes as the inverse problem becomes more difficult. In the Heat-1D difficulty sweep of \cref{fig:ablations}, both \xGD{} and \DIPG{} require the most iterations at the medium setting, reflecting increased curvature and noise. Across all regimes, however, \DIPG{} consistently converges in far fewer steps and shows the largest relative gain at the hard setting, roughly an order of magnitude fewer iterations than \xGD{}. This indicates that once the learned inverse stabilizes, \DIPG{} retains efficiency even as the forward problem becomes more nonlinear, whereas \xGD{} remains sensitive to scale and conditioning. The same figure also reports two JCP consistency tests and a qualitative recovery. Enabling JCP reduces the composition residual $\RJCP=\mathbb{E}_{\xi}\,\lVert J_g(f)J_f\,\xi-\xi\rVert^2$
 by several orders of magnitude, confirming near-inverse behavior of the learned maps and yielding fewer iterations under identical Armijo and projection rules. Together, these results show that lowering composition error directly translates into faster convergence.

Quantitatively, enabling JCP reduces $\RJCP$ by several orders of magnitude and shortens convergence. With JCP active on Heat-1D, the method reaches tolerance in about $2.6$ iterations with final RMSE $0.007$; disabling JCP raises the composition residual to $457.7$ and requires roughly $3.8$ iterations with higher error. Tying $V=W^\top$ removes degrees of freedom in the reverse map and degrades conditioning, while removing reconstruction and cycle terms does not harm convergence, indicating that the preconditioning effect is driven by the local inverse property rather than the auxiliary reconstruction losses.

We also track $\RJCP(x)$ during training as a runtime diagnostic. It decreases steadily as the reverse map stabilizes, aligning with validation error and indicating that improved composition $J_g(f)J_f$ yields better-scaled updates and faster convergence.

\section{Scalability and Discussion}

\begin{figure}[h]
\centering
\begin{adjustbox}{max width=0.36\linewidth}
\begin{tikzpicture}[
  scale=0.70,
  every node/.style={transform shape},
  small/.style={font=\scriptsize},
  tiny/.style={font=\scriptsize}
]

  \tikzset{
    >={Latex[length=1.8mm]},
    box/.style={draw, rounded corners=1pt, line width=0.6pt, inner sep=2pt, align=center, fill=#1},
    feat/.style={box=featcol},
    module/.style={box=modcol},
    op/.style={box=opcol},
    update/.style={box=updcol},
    flow/.style={->, line width=0.6pt, rounded corners=2pt, shorten >=0.5pt, shorten <=0.5pt}
  }

  \node[feat] (yup) at (-2.6,0.0) {\(\uparrow y\)};
  \node[feat] (rup) at ( 0.0,0.0) {\(\uparrow r_t\)};
  \node[feat] (xt)  at ( 2.6,0.0) {\(x_t\)};
  \node[tiny] at (0,0.48) {features};
  \node[tiny] at (0,0.85) {$\uparrow$: upsample to image grid};

  \node[op,     minimum width=6.0cm] (concat) at (0,-1.05)
        {concat \(\,[\uparrow y,\,\uparrow r_t,\,x_t]\,\)};
  \node[module, minimum width=6.6cm] (unet)   at (0,-2.30)
        {UNetSmall (3$\to$32$\to$1)\\predict \(\Delta x_t\)};

  \node[op] (gain)   at (-2.6,-3.35) {gain \(\alpha_t=\sigma(\gamma_t)\)};
  \node[op] (deltax) at ( 2.6,-3.35) {\(\Delta x_t\)};

  \node[op,     minimum width=4.2cm] (scale)  at (0,-4.30) {scale \(\alpha_t \cdot \Delta x_t\)};
  \node[op,     minimum width=5.1cm] (update) at (0,-5.10) {update \(x_t - \alpha_t \Delta x_t\)};
  \node[update, minimum width=3.4cm] (clip)   at (0,-5.85) {project \(\Pi_{[0,1]}\)};
  \node[feat]                 (xt1)           at (0,-6.55) {\(x_{t+1}\)};
  \node[tiny] at (0,-6.92) {next estimate};

  \draw[flow] (yup.south) -- (concat.north west);
  \draw[flow] (rup.south) -- (concat.north);
  \draw[flow] (xt.south)  -- (concat.north east);
  \draw[flow] (concat.south) -- (unet.north);
  \draw[flow] (unet.south west) .. controls +(-0.1,-0.7) and +(0,0.6) .. (gain.north);
  \draw[flow] (unet.south east) .. controls +( 0.1,-0.7) and +(0,0.6) .. (deltax.north);
  \draw[flow] (gain.south)   |- (scale.west);
  \draw[flow] (deltax.south) |- (scale.east);
  \draw[flow] (scale.south)  -- (update.north);
  \draw[flow] (update.south) -- (clip.north);
  \draw[flow] (clip.south)   -- (xt1.north);

  \coordinate (outerA) at ($(xt.east)+(1.5, 0.9)$);
  \coordinate (outerB) at ($(update.east)+(1.5, 0.8)$);
  \draw[flow] (xt.east) .. controls (outerA) and (outerB) .. (update.east);

  \node[tiny, anchor=west] at ($(xt1.east)+(0.8,0.0)$) {$\times\,N{=}6$ steps};
\end{tikzpicture}
\end{adjustbox}
\caption{DeceptronNet v0. A compact unrolled corrector using measurement and residual features.}
\label{fig:deceptronnet_v0}
\end{figure}
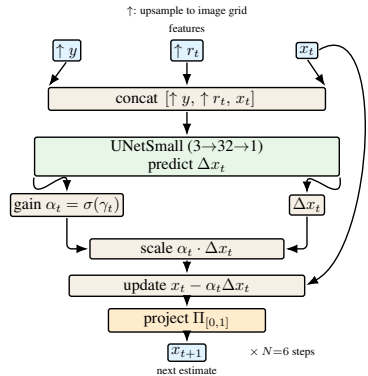

DeceptronNet v0 is a lightweight unrolled corrector that refines images in a few steps using measurement and residual features. At step $t$, inputs $F_t=[\,\uparrow y,\,\uparrow r_t,\,x_t\,]$ with $r_t=A_{\text{nom}}(x_t)-y$ pass through a compact U-Net (\texttt{UNetSmall}, $3{\to}32{\to}1$) predicting $\Delta x_t$. A learnable gain $\alpha_t=\sigma(\gamma_t)\in(0,1)$ scales the update, and the iterate advances as $x_{t+1}=\Pi_{[0,1]}\!\big(x_t-\alpha_t\Delta x_t\big)$. Depth is fixed at $N{=}6$, initialized with $x_0=\uparrow y$.

Training minimizes image error plus measurement consistency under $A_{\text{true}}$ (blur, mild nonlinearity, downsample, Poisson-like noise), without additional regularizers. For fairness, all methods share initialization, clamping, residual-based stopping ($0.3r_0$), iteration budget (80), and Armijo backtracking for the baseline solvers.

\begin{minipage}{\linewidth}\centering
\footnotesize\setlength{\tabcolsep}{4.5pt}
\captionof{table}{2D PSF results (mean over test set). Evaluated on $A_{\text{true}}$ with the same residual-based stopping rule. DNet reaches the target tolerance in a fixed, small number of learned steps, while LM and \xGD{} require many backtracked iterations.}
\label{tab:DNet_2d}
\begin{adjustbox}{max width=\linewidth}
\begin{tabular}{lcc}
\toprule
Method & Mean iterations to stop & Mean image RMSE \\
\midrule
LM (true model) & $69.25$ & $0.0883$ \\
x-GD (true model) & $80.00$ & $0.1271$ \\
DNet v0 (unrolled $N{=}6$) & $6.00$ & $0.0640$ \\
\bottomrule
\end{tabular}
\end{adjustbox}
\end{minipage}

DNet v0 converges rapidly under the same fairness conditions, reaching the error threshold in a small, fixed number of learned steps (Table~\ref{tab:DNet_2d}). This single-scale prototype demonstrates that amortized curvature and bounded updates can yield predictable convergence with minimal computation. While limited to simulated degradations, it forms the foundation for upcoming multi-scale and real-data variants. Its design emphasizes three aspects: (i) amortized curvature from residual features, (ii) stability through bounded gains and skip paths, and (iii) predictable compute from fixed iteration depth. Extending this framework to multi-scale operators, more realistic noise, and broader physical models is a natural next step.

\begin{figure}[H]
\centering
\begin{subfigure}{0.35\linewidth}
  \includegraphics[width=\linewidth]{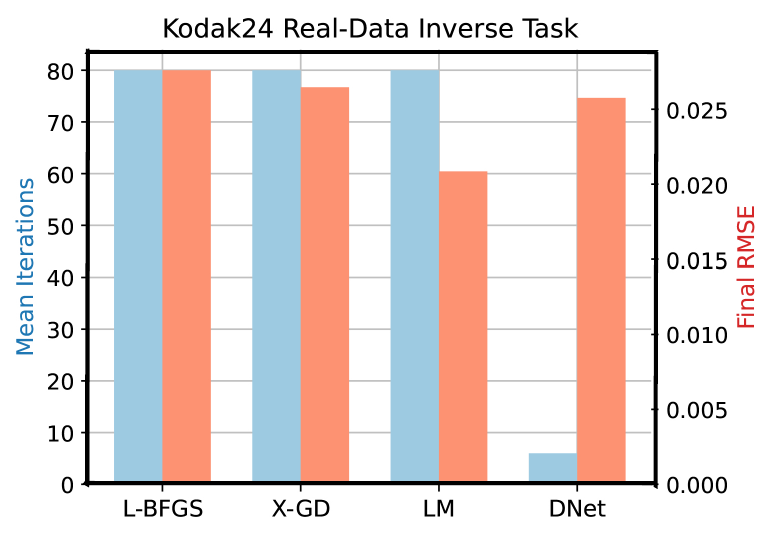}
  \caption{Hard real-image inverse task (Kodak24).}
  \label{fig:kodak24}
\end{subfigure}\hspace{8pt}
\begin{subfigure}{0.40\linewidth}
  \includegraphics[width=\linewidth]{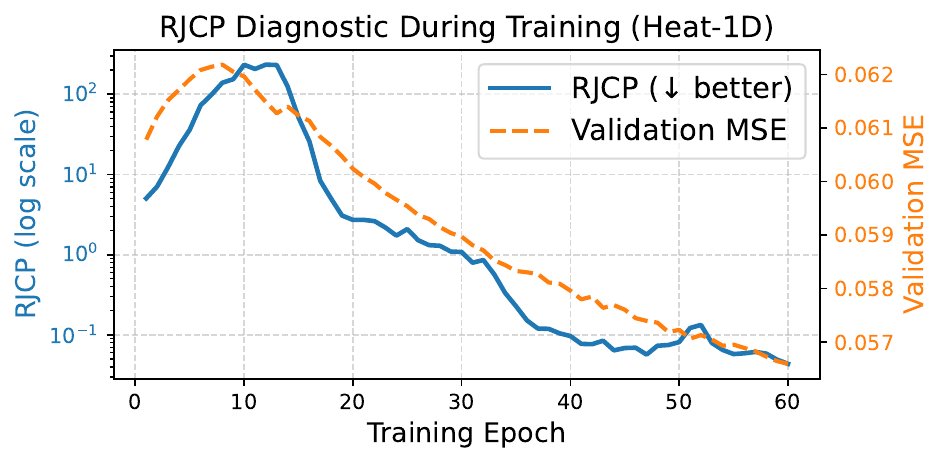}
  \caption{$\RJCP$ diagnostic evolution during training.}
  \label{fig:rjcp_diagnostic}
\end{subfigure}
\caption{Scalability and diagnostics. 
DNet remains stable under harder real-data settings (Kodak24), 
while D-IPG shows decreasing $\RJCP$ throughout training, indicating improved local invertibility.}

\label{fig:scalability_diagnostic}
\end{figure}

On the Kodak24 inverse task, all baselines use the same projection and Armijo backtracking. L-BFGS, LM, and \xGD{} converge slowly, while DNet reaches the residual threshold within six updates with competitive RMSE. Despite operating under noisy, downsampled conditions, the learned corrector maintains stability and reproducibility, showing that amortized updates generalize beyond synthetic settings. For the Deceptron (\DIPG{}), the $\RJCP$ diagnostic further reveals the mechanism behind this robustness. As training progresses, $\RJCP$ decreases steadily alongside validation error, confirming that the JCP indeed encourages the reverse map to act as a local left inverse rather than serving as a generic regularizer. This diagnostic correlates strongly with convergence speed and can be monitored at inference time to detect when surrogates fall outside their valid regime.

Our two prototypes, Deceptron and DeceptronNet (v0), highlight complementary philosophies. \DIPG{} enforces a learned local inverse through the Jacobian composition penalty, providing principled conditioning and interpretability, while DNet amortizes curvature through residual features, achieving fast, fixed-depth correction. Together they mark a first step toward lightweight, learned correctors for physical inverse problems. Faster, better-conditioned solvers can reduce compute cost and enable larger parameter sweeps in scientific pipelines such as imaging or system identification. However, misuse of learned surrogates outside their validity domain can yield overconfident reconstructions; we recommend explicit reporting of surrogate ranges and $\RJCP$ metrics in future learned-inverse work.

\section*{Code Availability}
All code, configuration files, and reproducibility notebooks for the Deceptron and DeceptronNet experiments are publicly available at:\\
\url{https://github.com/aadityakachhadiya/deceptron-ml4ps2025}

\section*{Acknowledgements}
We thank the ML4PS reviewers for their constructive feedback, which helped clarify several analyses. 
We also acknowledge open-source frameworks used for reproducibility, including PyTorch and NumPy.

\section*{Funding Statement}
This research received no external funding or institutional support. 
All computational experiments were performed independently by the author.

\section*{LLM Disclosure}
All core ideas, formulations, experiments, and writing structure were conceived and implemented by the author. 
Large Language Models (e.g., ChatGPT) were used for limited assistance in code debugging, LaTeX formatting, 
and minor language polishing. No model generated original research ideas or results.

\bibliographystyle{unsrtnat}
\bibliography{references}

\clearpage
\appendix
\section{Extra Results and Supporting Algorithms}

This appendix compiles the supporting algorithms, diagnostic plots, and theoretical derivations complementing the main text. While not essential for reproducing the core experiments, these details clarify implementation, interpretability, and the underlying optimization behavior of the proposed methods.

\subsection{Optimization Algorithms}

All solvers share the same fairness protocol: identical projection $\Pi_{\mathcal C}$, Armijo parameter $c{=}10^{-4}$ (up to eight halvings), relaxation $\rho{=}0.4$, and identical initialization and stopping rules.  
The following pseudocode specifies each update exactly as used in experiments.

\begin{algorithm}[h]
\caption{D-IPG (Deceptron Inverse-Preconditioned Gradient) with shared Armijo and relaxation}
\label{alg:dipg}
\begin{algorithmic}[1]
\STATE \textbf{Inputs:} surrogate $f_W$, reverse $g_V$, projector $\Pi_{\mathcal C}$, target $y^\ast$, init $x_0$, step $\alpha_0$, relaxation $\rho$, Armijo $c$, halvings $H$, tolerance $\varepsilon$
\FOR{$t=0,1,\dots$}
  \STATE $y_t \gets f_W(x_t)$; $r_t \gets y_t - y^\ast$; $\Phi_t \gets \tfrac12\|r_t\|^2$
  \STATE compute $\nabla\Phi(x_t)$ by reverse-mode AD
  \STATE $\alpha \gets \alpha_0$; accepted $\gets$ \textbf{false}
  \FOR{$h=0,\dots,H$}
    \STATE $y_{\text{prop}} \gets y_t - \alpha\, r_t$; $x_{\text{prop}} \gets g_V(y_{\text{prop}})$
    \STATE $p \gets x_{\text{prop}} - x_t$
    \STATE $x_{\text{trial}} \gets \Pi_{\mathcal C}\big((1{-}\rho)x_t + \rho(x_t{+}p)\big)$
    \STATE $\Phi_{\text{trial}} \gets \tfrac12\|f_W(x_{\text{trial}})-y^\ast\|^2$; $g^\top p \gets \nabla\Phi(x_t)^\top p$
    \IF{$\Phi_{\text{trial}} \le \Phi_t + c\,\rho\,g^\top p$}
      \STATE $x_{t+1}\gets x_{\text{trial}}$; accepted $\gets$ \textbf{true}
      \STATE \textbf{break}
    \ELSE
      \STATE $\alpha \gets \alpha/2$
    \ENDIF
  \ENDFOR
  \IF{\textbf{not} accepted}
    \STATE \textbf{break}
  \ENDIF
  \STATE stop if normalized residual $\le \varepsilon$
\ENDFOR
\end{algorithmic}
\end{algorithm}

\begin{algorithm}[h]
\caption{x-GD (projected gradient) with shared Armijo and relaxation}
\label{alg:xgd}
\begin{algorithmic}[1]
\STATE \textbf{Inputs:} $f_W$, $\Pi_{\mathcal C}$, $y^\ast$, $x_0$, step $\eta_0$, $\rho$, $c$, $H$, $\varepsilon$
\FOR{$t=0,1,\dots$}
  \STATE compute $\nabla\Phi(x_t)$ for $\Phi(x)=\tfrac12\|f_W(x)-y^\ast\|^2$; set $\eta\gets\eta_0$
  \FOR{$h=0,\dots,H$}
    \STATE $p \gets -\eta\,\nabla\Phi(x_t)$
    \STATE $x_{\text{trial}}\gets\Pi_{\mathcal C}\big((1{-}\rho)x_t + \rho(x_t{+}p)\big)$
    \IF{$\Phi(x_{\text{trial}}) \le \Phi(x_t) + c\,\rho\,\nabla\Phi(x_t)^\top p$}
      \STATE $x_{t+1}\gets x_{\text{trial}}$
      \STATE \textbf{break}
    \ELSE
      \STATE $\eta \gets \eta/2$
    \ENDIF
  \ENDFOR
  \STATE stop if normalized residual $\le \varepsilon$
\ENDFOR
\end{algorithmic}
\end{algorithm}

\begin{algorithm}[h]
\caption{GN/LM (damped Gauss--Newton with CG) under shared projector, relaxation, and Armijo}
\label{alg:gn}
\begin{algorithmic}[1]
\STATE \textbf{Inputs:} $f_W$, $\Pi_{\mathcal C}$, $y^\ast$, $x_0$, step $\alpha_0$, $\rho$, $c$, $H$, damping $\lambda$, CG iters $K$
\FOR{$t=0,1,\dots$}
  \STATE $r \gets f_W(x_t) - y^\ast$; $b \gets J^\top r$ via VJP; define $v\mapsto (J^\top J + \lambda I)v$ via JVP+VJP
  \STATE solve $(J^\top J + \lambda I)\Delta x = -b$ by $K$ CG iterations
  \STATE $\alpha \gets \alpha_0$
  \FOR{$h=0,\dots,H$}
    \STATE $x_{\text{trial}}\gets\Pi_{\mathcal C}\big((1{-}\rho)x_t + \rho(x_t{+}\alpha\Delta x)\big)$
    \IF{$\Phi(x_{\text{trial}}) \le \Phi(x_t) + c\,\rho\,\nabla\Phi(x_t)^\top (\alpha\Delta x)$}
      \STATE $x_{t+1}\gets x_{\text{trial}}$
      \STATE \textbf{break}
    \ELSE
      \STATE $\alpha \gets \alpha/2$
    \ENDIF
  \ENDFOR
\ENDFOR
\end{algorithmic}
\end{algorithm}

\begin{algorithm}[h]
\caption{DeceptronNet v0 (single-scale; unrolled $N{=}6$)}
\label{alg:DNet}
\begin{algorithmic}[1]
\STATE \textbf{Inputs:} measurement $y$, nominal $A_{\text{nom}}$, current $x_t$, gain $\alpha_t=\sigma(\gamma_t)$
\STATE $r_t \gets A_{\text{nom}}(x_t)-y$; $F_t \gets [\,\uparrow y,\,\uparrow r_t,\,x_t\,]$
\STATE $\Delta x_t \gets \texttt{UNetSmall}(F_t)$
\STATE $x_{t+1} \gets \mathrm{clip}_{[0,1]}(x_t - \alpha_t\,\Delta x_t)$
\end{algorithmic}
\end{algorithm}

\section{Extra Results}

\begin{figure}[h]
\centering
{\includegraphics[width=0.31\linewidth]{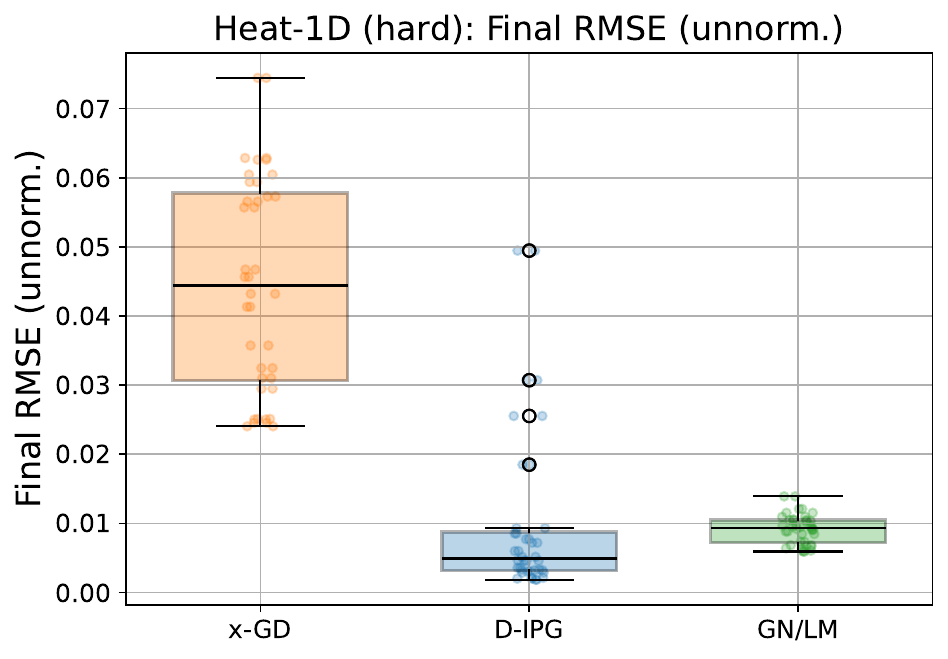}}
\hfill
{\includegraphics[width=0.31\linewidth]{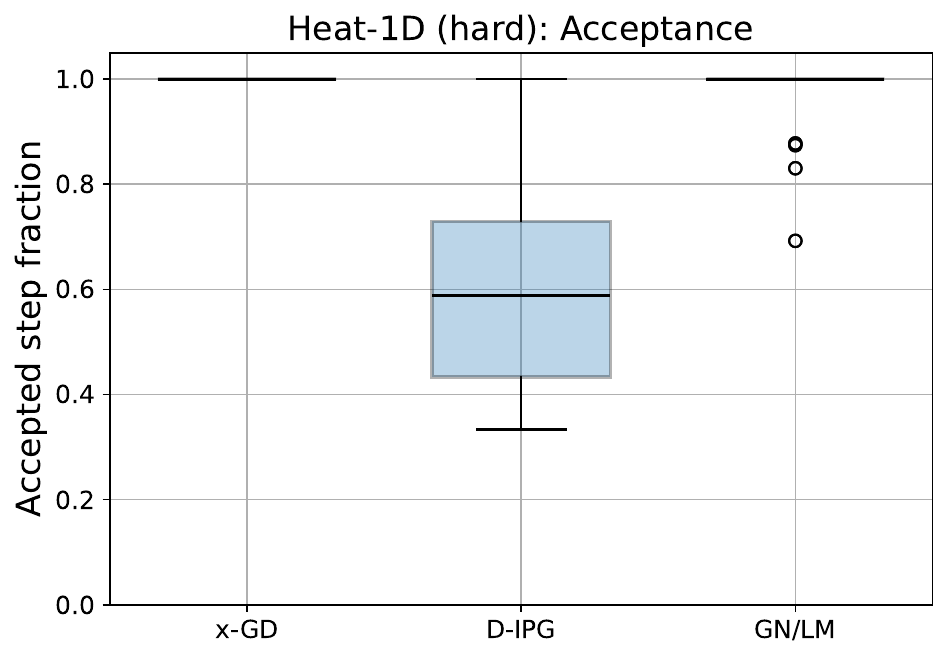}}
\hfill
{\includegraphics[width=0.31\linewidth]{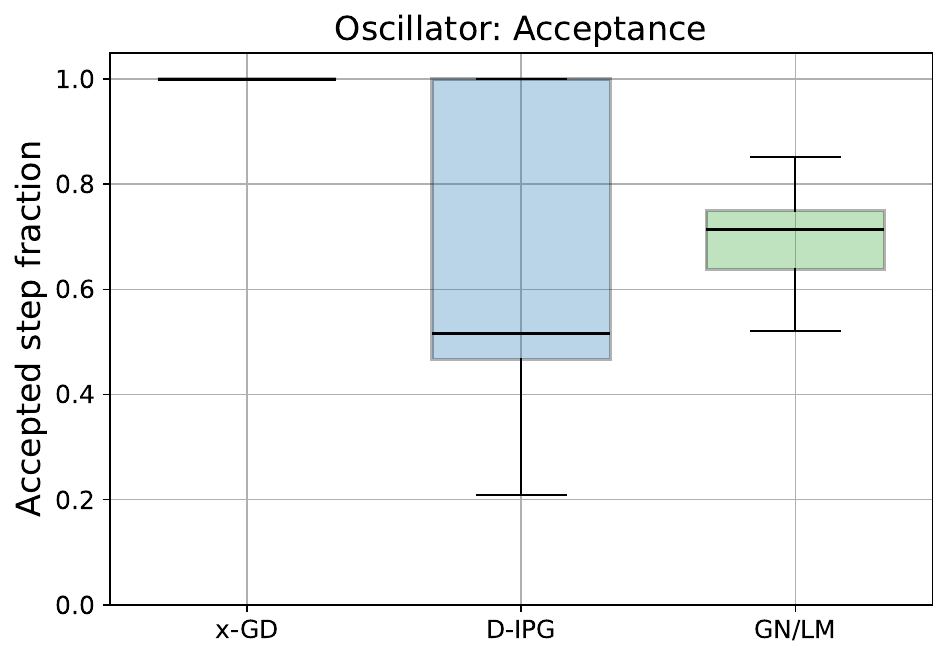}}
\caption{Additional box plots. Left: Heat-1D RMSE (unnormalized). Middle/Right: acceptance ratios for Heat-1D and Oscillator. Note that acc$<1$ reflects larger proposed moves under the shared Armijo rule and does \emph{not} indicate instability.}
\label{fig:app-boxes}
\end{figure}

\begin{table}[h]
\caption{Ablation study under {\xGD{} vs \DIPG{}}. The JCP term enhances local conditioning and convergence speed without affecting runtime cost.}
\centering
\small
\setlength{\tabcolsep}{5pt}
\begin{tabular}{l l c c c}
\toprule
Ablation & Method & Iters (mean$\pm$std) & Final RMSE & Acceptance \\
\midrule
$-$JCP          & \xGD  & 56.3$\pm$37.0  & 0.0436  & 1.000 \\
                & \DIPG & 3.25$\pm$1.18  & 0.0159  & 0.606 \\
$V=W^\top$      & \xGD  & 106.9$\pm$41.3 & 0.0444  & 1.000 \\
                & \DIPG & 16.2$\pm$8.55  & 0.0894  & 0.061 \\
$-$rec/$-$cycle & \xGD  & 44.8$\pm$21.9  & 0.0361  & 1.000 \\
                & \DIPG & 2.60$\pm$0.92  & 0.0086  & 0.595 \\
$-$comp         & \xGD  & 44.7$\pm$21.9  & 0.0365  & 1.000 \\
                & \DIPG & 2.60$\pm$0.92  & 0.0070  & 0.570 \\
\bottomrule
\end{tabular}
\label{tab:app-ablation}
\end{table}

\begin{table}[h]
\caption{Core per-iteration operations (excluding shared backtracking $f$-evaluations). The Jacobian Composition Penalty (JCP) acts only at training time and adds no runtime cost.}
\centering
\small
\setlength{\tabcolsep}{6pt}
\begin{tabular}{l l}
\toprule
Method & Operations per iteration \\
\midrule
x-GD   & one reverse-mode gradient $\nabla_x \Phi$ \\
D-IPG  & one reverse-mode grad (for Armijo) $+$ $f$ $+$ $g$ \\
GN/LM  & solve $(J^\top J+\lambda I)\Delta x=-J^\top r$ by CG; each CG $\approx$ 1 JVP $+$ 1 VJP \\
\bottomrule
\end{tabular}
\label{tab:app-ops}
\end{table}

\begin{minipage}{\linewidth}\centering
\footnotesize\setlength{\tabcolsep}{4.5pt}
\captionof{table}{Kodak24 inverse task. Quantitative RMSE, iteration count, and mean wall-time per sample. The $\sigma$-Mismatch test (train $\sigma{=}4.0$, eval $\sigma{=}3.0$) probes robustness under surrogate–data noise shift. DNet maintains low error and fixed-step convergence.}
\label{tab:kodak24}
\begin{adjustbox}{max width=\linewidth}
\begin{tabular}{lcccc}
\toprule
Setting & Method & RMSE & Iters & Time (s) \\
\midrule
\multirow{4}{*}{Normal ($\sigma{=}3.0$)} 
 & L-BFGS & 0.0276 & 80 & 0.173 \\
 & X-GD   & 0.0265 & 80 & 0.0054 \\
 & LM     & 0.0209 & 80 & 0.0058 \\
 & DNet   & 0.0258 & 6  & 0.0067 \\
\midrule
\multirow{4}{*}{Hard ($\sigma{=}4.0$)} 
 & L-BFGS & 0.0604 & 100 & 0.181 \\
 & X-GD   & 0.0589 & 100 & 0.0063 \\
 & LM     & 0.0550 & 100 & 0.0064 \\
 & DNet   & 0.0575 & 6   & 0.0046 \\
\midrule
\multirow{4}{*}{$\sigma$-Mismatch (train 4.0, eval 3.0)} 
 & L-BFGS & 0.0548 & 100 & 0.180 \\
 & X-GD   & 0.0535 & 100 & 0.0063 \\
 & LM     & 0.0488 & 100 & 0.0062 \\
 & DNet   & 0.0525 & 6   & 0.0045 \\
\bottomrule
\end{tabular}
\end{adjustbox}
\end{minipage}

Here $\sigma$ denotes the surrogate’s assumed observation-noise level during training and evaluation. 
A higher $\sigma$ corresponds to a noisier forward model used for generating synthetic measurements.

\subsection{Relation to Gauss--Newton}

The classical Gauss--Newton (GN) method solves least-squares inverse problems using
$d_{\mathrm{GN}}=-\,(J^\top J+\lambda I)^{-1}J^\top r$, 
where $J$ is the Jacobian of the forward model.
Our D-IPG update, $d_{\mathrm{D\text{-}IPG}}=-\alpha\,B^{-1}J^\top r$, 
replaces this curvature matrix with a learned preconditioner $B$.
When $B$ approximates $J^\top J+\lambda I$, the two directions become nearly identical,
and convergence behavior matches that of GN up to a small scaling factor.
The Jacobian Composition Penalty (JCP) encourages this alignment by enforcing
$J_g(f(x))J_f(x)\!\approx\!I$,
and the runtime diagnostic $\mathrm{RJCP}$ measures how close this condition holds.
Lower $\mathrm{RJCP}$ therefore indicates stronger curvature alignment
and faster, more stable optimization.

\paragraph{Deviation from GN (local, range-restricted).}
Let $J = J_f(x_t)$ denote the Jacobian of the forward map at the current iterate.
Assume $J$ has full column rank in a neighborhood of the solution.
We write the learned reverse Jacobian as $J_g(f(x_t)) = J^{+} + E_g$,
where $J^{+} = (J^{\top}J)^{-1}J^{\top}$ is the Moore--Penrose pseudoinverse
and $E_g$ represents the residual error of the learned local inverse.
For the component of the residual $r$ that lies in $\mathrm{range}(J)$
(i.e., near a solution where $r \approx J u$ for some $u$),
the D\mbox{-}IPG and Gauss--Newton steps are approximately
$\Delta x_{\mathrm{dipg}} = -\alpha\,J_g(f(x_t))\,r$
and $\Delta x_{\mathrm{GN}} = -\,J^{+} r$, respectively.
Their difference satisfies
\[
\boxed{\;
\|\Delta x_{\mathrm{dipg}} - \Delta x_{\mathrm{GN}}\|
\;\le\;
\alpha\,\frac{\|J_gJ - I\|_2}{\sigma_{\min}(J)}\,\|r\|
\;}
\]
where $\|\cdot\|_2$ denotes the spectral norm
and $\sigma_{\min}(J)$ is the smallest singular value of $J$.
Thus, as the JCP target $\|J_gJ - I\|_2 \!\to\! 0$,
the D\mbox{-}IPG direction converges to the Gauss--Newton direction
up to the scalar step size~$\alpha$.
The Jacobian Composition Penalty (JCP) enforces this alignment during training
but incurs no additional runtime cost during inference.

\section{Jacobian Composition Penalty: Diagnostics}
\label{app:rjcp}

Recall the runtime Jacobian composition error ($\RJCP$) from the main text,
\[
\RJCP(x)=\E_{\xi}\|J_g(f(x))J_f(x)\xi-\xi\|^2,
\]
an unbiased estimator of $\|J_g(f(x))J_f(x)-I\|_F^2$ via Hutchinson’s identity. $\RJCP$ measures how well the learned reverse map $g$ acts as a local left inverse of $f$: $\RJCP(x)=0$ if and only if $J_g(f(x))J_f(x)=I$. Lower values indicate near-unit scaling and low cross-coupling, corresponding to well-conditioned updates, while larger values signal mis-scaling or axis mixing.

Computation is efficient: with $k$ probes $\xi_j$, compute $v_j=\mathrm{JVP}_{f_W}(x;\xi_j)$, then $u_j=\mathrm{JVP}_g(y;v_j)$ at $y=f_W(x)$, and accumulate $\|u_j-\xi_j\|^2$. Averaging over $j=1..k$ yields $\RJCP(x)$. This requires only JVP/VJP products, no explicit Jacobians, and costs $\mathcal{O}(k)$ forward/adjoint passes. In practice, $k=2$–4 probes suffice.

During training, $\RJCP$ also appears as a weighted penalty (JCP) to shape $g$ toward acting as a left inverse. Reductions in $\RJCP$ across epochs correlate with fewer iterations to reach tolerance, while plateaus at high values typically reflect an unstable surrogate, an overly strong or early JCP weight, or severe non-identifiability. At evaluation, we log $\RJCP$ as a scalar diagnostic and observe that monotone decreases track improved iteration counts.

Interpretation is straightforward: low $\RJCP$ corresponds to well-scaled, stable steps; moderate values to partial conditioning; and high values to weak or unstable preconditioning. $\RJCP$ is diagnostic only and does not imply global invertibility, but it is informative within the surrogate’s validity region. In highly non-identifiable regimes (multiple $x$ yielding similar $y$), $\RJCP$ cannot resolve global ambiguity but still reflects local conditioning.

Implementation is lightweight: normalize outputs for the objective and line search, warm up the JCP weight after the forward fit stabilizes, use $k=2$–4 Rademacher probes per batch, apply only JVP/VJP products, keep the spectral term on $W^\top W$ modest, avoid tying $V=W^\top$, and apply Armijo/backtracking consistently with baselines. If $\RJCP$ remains high, delay JCP warm-up, improve the surrogate near initialization, modestly tune $\alpha$ or $\lambda_{\text{JCP}}$, increase probes slightly, or switch to projected $\RJCP$ in under-determined regimes.

\section{Limitations of Deceptron}
\label{app:limits}

Deceptron relies on a reasonably accurate surrogate model, locality of linearization, and identifiable structure; outside these regimes its benefits diminish. 
If the surrogate is poorly fitted or strongly rank-deficient, corrective updates become unstable and $\RJCP$ remains persistently high. 
In highly non-identifiable problems, where many solutions map to similar measurements, the method cannot resolve global ambiguity and only improves conditioning locally.

The approach further assumes that the constraint projector preserves most of the proposed update; if projections dominate, effective progress is lost. 
Performance is also sensitive to the scheduling of step-size gains $\alpha_t$ and to the timing and weighting of the JCP term, which may require manual tuning for stability. 
The deliberately lightweight network improves efficiency but limits expressivity, making the method less competitive in problems that require strong nonlinearities, long-range dependencies, or nonlocal priors. 
In such settings, richer architectures may achieve higher fidelity at the cost of speed.

Finally, Deceptron is intended as a corrective accelerator rather than a full solver replacement, with the largest gains when the surrogate provides a useful local model. 
This motivates the \textbf{DeceptronNet variant for 2D (and higher-dimensional)} tasks, where a slightly richer architecture and multi-scale design help overcome some of these limitations. 
While still lightweight, DeceptronNet demonstrates improved stability on real datasets such as Kodak24 and better captures spatial structure and nonlocal correlations, extending the practical range of our approach beyond the single-scale version.

\end{document}